\documentclass[letterpaper]{article} % DO NOT CHANGE THIS
\usepackage{aaai2027}
\usepackage[hyphens]{url}  % DO NOT CHANGE THIS
\usepackage{graphicx} % DO NOT CHANGE THIS
\usepackage{natbib}  % DO NOT CHANGE THIS AND DO NOT ADD ANY OPTIONS TO IT
\usepackage{caption} % DO NOT CHANGE THIS AND DO NOT ADD ANY OPTIONS TO IT
\usepackage{booktabs}
\usepackage{amsmath,amssymb}
\usepackage{enumitem}

\newcommand{\R}{\mathbb{R}}
\newcommand{\E}{\mathbb{E}}
\newcommand{\cL}{\mathcal{L}}

\title{FabriVLA: A Lightweight Vision-Language-Action Model with Conformal Action Chunk Uncertainty}
\author{
    Shiyuan Yang$^{1,}\thanks{These authors contributed equally.}$,
    Borong Zhang$^{1,*}$,
    Jizheng Zhang$^{1,*}$,
    Zhijia Tao$^{1,*}$, \\
    Junfei Guo$^{2}$, 
    Donglai Ran$^{3}$,
    Xu Bian$^{3,\dagger}$,
    Qingbiao Li$^{1,}\thanks{Corresponding Author.}$
  }
\affiliations{
    $^{1}$University of Macau\\
    $^{2}$Mese Technology Limited Co., Ltd.,
    $^{3}$FabriX team at Youibot Robotics Co., Ltd.
}
\begin{document}

\maketitle
% ============================================================================
% ABSTRACT
% ============================================================================
\begin{abstract}
Vision-Language-Action (VLA) models have become a leading paradigm for general purpose robotic manipulation, but their computational cost and limited uncertainty awareness hinder practical deployment. We present FabriVLA, a lightweight VLA that fuses shallow and intermediate VLM layers to preserve fine-grained visual features, and gates self-attention among action tokens so that its flow matching head admits inter step structure only as far as training warrants. Trained end-to-end in a single stage, FabriVLA reaches a state-of-the-art 90.0\% average success on Meta-World MT50 with only 0.88B parameters. We further introduce Joint Conformal Action Chunk Calibration (JCAC), a post-training method that augments a frozen policy with a lightweight residual scale head. From a single policy query, JCAC turns a learned elementwise error scale into a set that covers the whole executed action prefix at a user chosen confidence level, 3.3$\times$ tighter in mean radius than an unconditional conformal set. On LIBERO-Safety, these bounds rank rollouts by risk before execution, supporting risk ranked review. Together, FabriVLA and JCAC provide a lightweight and auditable framework for multi-task manipulation with calibrated action uncertainty.
\end{abstract}

% ============================================================================
% 1. INTRODUCTION
% ============================================================================
\section{Introduction}

Vision-Language-Action (VLA) models have emerged as a promising paradigm for generalist robot manipulation, leveraging pretrained vision-language models (VLMs) to ground language instructions in visual observations and produce executable action sequences \citep{pmlr-v229-zitkovich23a,kim2024openvla,black2024pi0}. While larg scale VLAs with tens of billions of parameters achieve impressive results, their computational cost and inference latency pose practical challenges for real time robotic control. This motivates the development of lightweight VLA architectures that balance performance with efficiency. Efficiency alone, however, is not sufficient for deployment: a policy that acts fast but silently gives an operator no indication of when its predicted actions should not be trusted. This raises a question that is essential yet rarely discussed in the VLA literature. How can a compact VLA achieve strong manipulation performance and, at the same time, expose how much each of its own action chunks can be trusted before that chunk is executed?

\begin{figure}[t]
\centering
\includegraphics[width=\columnwidth]{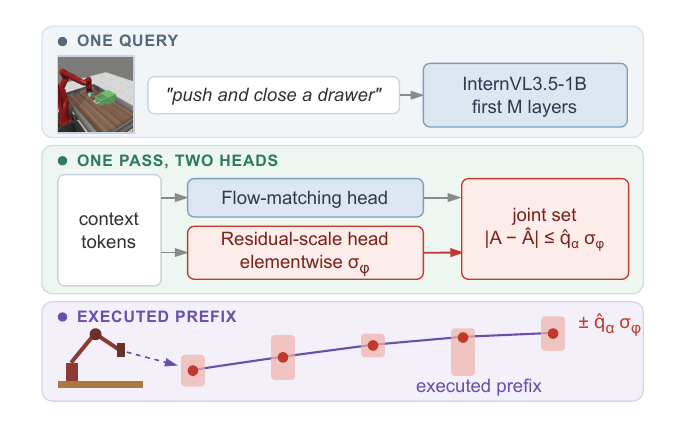}
\caption{One policy query yields both an action chunk and an elementwise error scale; a single conformal quantile then covers every entry of the executed prefix at once. The JCAC head adds 0.58M parameters to the policy and attains a 3.3$\times$ smaller mean radius than a global conformal radius at the same 90\% joint prefix target.}
\label{fig:teaser}
\end{figure}

To answer this question, we propose \textbf{FabriVLA}, a lightweight VLA model that achieves competitive performance both on simulation benchmarks and in real-world experiments while requiring only a fraction of the parameters used by prevailing VLA models \citep{kim2024openvla,black2024pi0,yan2025robotron}, together with \textbf{JCAC}, a post training calibration method that makes the action chunks of a frozen policy auditable before they are executed (Figure~\ref{fig:teaser}). Trained end-to-end in a single stage, FabriVLA reaches a 90.0\% success rate on Meta-World MT50 with 0.88B parameters (Table~\ref{tab:comparison}). The main contributions are summarized.

% Gated self-attention lets action tokens within the prediction horizon attend to each other through a learnable gate initialized to zero, so each block starts out as a cross-attention-only transformer and admits inter-step dependencies only as far as training warrants. Shallow VLM layer fusion combines the final VLM layer with an intermediate one, exposing both semantic context and lower-level spatial detail to the action head, which is important for precise object localization and contact rich manipulation.

\begin{itemize}[leftmargin=*,itemsep=2pt,topsep=3pt]
\item A lightweight VLA architecture that preserves manipulation precision, so a compact model need not trade accuracy for size.
\item JCAC, a post-training layer that makes a frozen policy auditable: one query yields both the action chunk and an elementwise error scale, and a single conformal quantile turns that scale into a distribution free set covering the whole executed prefix at once.
\item Evidence that the two combine: a 0.88B parameter model trained in one stage without robot data pretraining leads the compared VLAs on MT50, and the same single stage recipe carries over without modification to a real arm, where the calibrated policy runs in closed loop and its deployment time scores are logged.
\end{itemize}

% ============================================================================
% 2. RELATED WORK
% ============================================================================
\section{Related Work}

\paragraph{Lightweight VLA Architectures.}
VLA models map multimodal observations and language instructions to robot actions, as exemplified by RT-2~\citep{pmlr-v229-zitkovich23a} and OpenVLA~\citep{kim2024openvla}. Recent designs span different parameter and pretraining regimes. TinyVLA (1.3B)~\citep{wen2024tinyvla}, SmolVLA (2.3B)~\citep{shukor2025smolvla}, RoboTron-Mani (4B)~\citep{yan2025robotron}, Evo-1 (0.8B)~\citep{lin2025evo1}, and Evo-Depth (0.9B)~\citep{evodepth2026} are reported without robot data pretraining, whereas $\pi_0$ (3.5B)~\citep{black2024pi0} uses robot data pretraining and LA4VLA (1B)~\citep{lin2026la4vla} uses mixed pretraining. What these designs share is that the compact policy is delivered as a black box: none of them exposes, at deployment time, how far a produced action chunk may be from a correct one. Within this spectrum, FabriVLA uses 0.88B parameters and no separate robot data pretraining stage, and pairs that budget with an explicit deployment time uncertainty interface.

\paragraph{Generative Action Policies and Hierarchical Features.}
Predicting a chunk of future actions rather than one step at a time is now standard in imitation learning~\citep{zhao2023act}, and generative robot policies model such chunks using diffusion~\citep{chi2023diffusion} or flow matching~\citep{lipman2023flow}, with flow based action generation also adopted by recent VLAs~\citep{black2024pi0,lin2025evo1}. Such policies must represent inter step structure while retaining the spatial detail required for precise manipulation. Hypercolumns, feature pyramids, and transformer based dense predictors show that combining intermediate and final representations can preserve fine spatial structure together with high level semantics~\citep{hariharan2015hypercolumns,lin2017fpn,ranftl2021vision}. Rather than relying on final layer conditioning and cross-attention alone, FabriVLA fuses intermediate and final layer VLM features and introduces zero initialized gated self-attention among action tokens, enabling its flow matching head to model spatial context and inter step dependencies within each generated chunk.

\paragraph{Action Uncertainty and Conformal Calibration.}
Uncertainty in learned robot policies has been estimated through ensemble disagreement in imitation learning~\citep{menda2019ensembledagger}, which buys a signal at the cost of several forward passes per decision. Conformal prediction instead provides finite-sample, distribution-free marginal coverage under exchangeability~\citep{vovk2005algorithmic,angelopoulos2021conformal}, with extensions for input-dependent uncertainty~\citep{romano2019cqr} and applications to sequential decision making~\citep{ren2023knowno,lindemann2023safe,dixit2023adaptive,sun2024copula}. Among VLA methods, ReconVLA is the closest comparator: it scores multiple stochastic action candidates and selects the candidate predicted to be most reliable~\citep{chen2026reconvla}. In contrast, JCAC requires only one policy query and constructs a simultaneous elementwise prediction set over the entire action prefix executed before replanning, jointly calibrating errors across all executed steps and action dimensions. A complementary line asks whether the current observation resembles the training distribution at all, through feature-space nearest-neighbour scores~\citep{sun2022knnood} and task-driven out-of-distribution tests with statistical guarantees for robot learning~\citep{farid2022taskdriven}. That question is distinct from the action-residual question JCAC answers, and Section~\ref{sec:real_robot} reports a nearest-neighbour context diagnostic of this kind alongside, not in place of, the calibrated action set.

% ============================================================================
% 3. METHOD
% ============================================================================
\section{Method}

\subsection{Problem Formulation}

We consider language conditioned multi-task manipulation. At decision step $k$ the agent receives an RGB observation $\mathbf{o}_k$, a proprioceptive state $\mathbf{s}_k \in \R^{d_s}$, and a task instruction $\ell$ fixed within an episode. A policy $\pi_\theta$ maps $(\mathbf{o}_k, \mathbf{s}_k, \ell)$ to an action chunk $\widehat{\mathbf{A}}_k \in \R^{H \times D}$ covering the next $H$ control steps, where actions are padded to a common width $D$ so that one head can serve embodiments with different actuation counts and the current embodiment's $d$ valid dimensions are read out at execution time. Execution is receding horizon: only the prefix $\widehat{\mathbf{A}}^e_k = \widehat{\mathbf{A}}_{k,1:H_e,1:d}$ with $H_e \ll H$ reaches the robot before the next query, and the remaining steps are discarded. The executed prefix, not the full chunk, is therefore what both the policy and its calibration layer must get right. We reserve $k$ for decision steps and $t \in [0,1]$ for the flow matching time variable.

\begin{figure*}[t]
\centering
\includegraphics[width=0.98\textwidth]{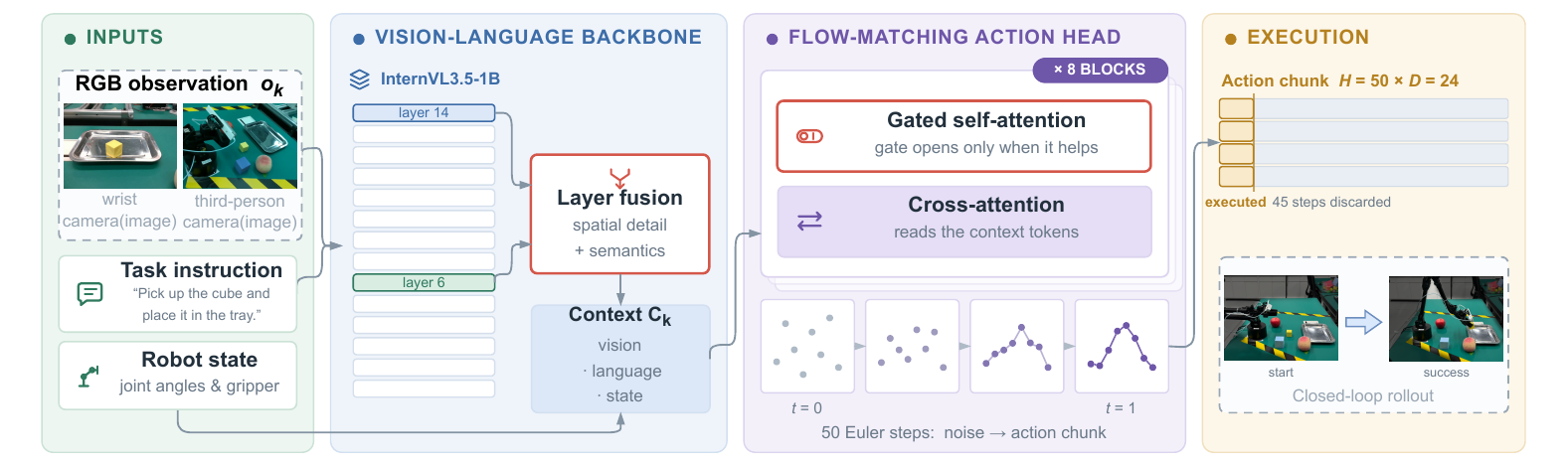}
\caption{Overview of FabriVLA. RGB views, the task instruction, and the robot state are encoded by a truncated InternVL3.5-1B backbone, with a shallow layer tapped and fused into the deep layer by a learned linear projection so that the context tokens carry spatial detail as well as task semantics. The flow matching head stacks blocks of gated self-attention over the action tokens followed by cross-attention to that context; integrating its velocity field carries noise to a complete action chunk.}
\label{fig:architecture}
\end{figure*}

\subsection{Backbone and Encoders}

Figure~\ref{fig:architecture} gives the overall architecture. We adopt InternVL3.5 \citep{wang2025internvl35} as the vision-language backbone, retaining only the first $M$ layers of its language model and leaving the vision encoder intact. The remaining interfaces follow standard VLA practice: images and instruction are consumed as a single interleaved sequence, from which the backbone emits contextualized tokens $\mathbf{C}^{(M)}_k \in \R^{N_c \times e}$ at its last retained layer, which Section~\ref{sec:fusion} fuses with a shallow layer into the context $\mathbf{C}_k$ that the action head consumes; the proprioceptive state $\mathbf{s}_k$, zero padded like the actions, is appended as one further token of width $e$; and the noisy action chunk $\mathbf{x}_t \in \R^{H \times D}$ is lifted to width $e$ by a multi-embodiment action encoder. Encoder depths and widths are given in Appendix~A of the supplementary material, to which all later appendix pointers refer.

\subsection{Gated Action Self-Attention}

Steps within a chunk are far from independent: consecutive actions share a direction of motion and a contact phase, which a head that predicts them in isolation must rediscover through the context tokens alone. Self-attention over action tokens is the natural remedy, but enabling it from the first gradient step is counterproductive, since the randomly initialized head would then mix uninformative tokens into exactly the representations that cross-attention is trying to ground. We therefore gate it.

The head is a stack of $L$ identical blocks operating on action tokens $\mathbf{A} \in \R^{H \times e}$ and context tokens $\mathbf{C} \in \R^{N_c \times e}$. Each block first applies multi-head self-attention over the action tokens through a learnable scalar gate $\lambda \in \R$ initialized to zero,
\begin{equation}
    \mathbf{A}' = \mathbf{A} + \lambda \cdot \text{SelfAttn}\big(\text{LayerNorm}(\mathbf{A})\big),
\end{equation}
so that the block starts out exactly as a cross-attention block and departs from it only as far as the gradient on $\lambda$ warrants. The action tokens then attend to the context,
\begin{equation}
    \mathbf{A}'' = \mathbf{A}' + \text{CrossAttn}\big(\text{LayerNorm}(\mathbf{A}'),\; \mathbf{C},\; \mathbf{C}\big),
\end{equation}
and a feed forward network conditioned on the flow timestep closes the block,
\begin{equation}
    \mathbf{A}''' = \mathbf{A}'' + \text{FFN}\big(\text{LayerNorm}(\mathbf{A}'') + \text{TimeEmb}(t)\big),
\end{equation}
where $\text{TimeEmb}(t) \in \R^{e}$ is a sinusoidal encoding of the flow matching timestep (Appendix~A).

The output of the final block is layer normalized, flattened over the horizon, projected back to width $e$ by a learned pooling projection $\R^{He} \to \R^{e}$, and decoded by a two layer MLP into the predicted velocity field $\mathbf{v}_\theta(\mathbf{x}_t, t \mid \mathbf{s}_k, \mathbf{o}_k, \ell) \in \R^{H \times D}$, where $\theta$ collects all trainable parameters of the backbone and the head. We abbreviate this as $\mathbf{v}_\theta(\mathbf{x}_t, t \mid \cdot)$ whenever the conditioning is clear from context. The pooling projection is a full linear map over all $H \times e$ entries rather than an average over steps, so per step information survives it.

\subsection{Shallow VLM Layer Fusion}
\label{sec:fusion}

The last retained backbone layer carries the most abstract semantics, but abstraction discards what contact rich manipulation depends on: object boundaries, relative offsets, and the pixel support of the target, all of which intermediate layers still hold. FabriVLA therefore conditions the action head on a fusion of a deep and a shallow layer rather than on the final layer alone. Let $\mathbf{C}^{(M)}$ and $\mathbf{C}^{(m)}$ with $m < M$ denote the token sequences at the final and at an intermediate retained layer. We concatenate them along the feature dimension and project back to the token width,
\begin{equation}
    \mathbf{C} = \big[\mathbf{C}^{(M)} \;\big|\; \mathbf{C}^{(m)}\big]\,\mathbf{W},
    \qquad \mathbf{W} \in \R^{2e \times e},
\end{equation}
initializing $\mathbf{W} = [\mathbf{I} \mid \mathbf{0}]^{\!\top}$ so that the fused context is identical to the deep only context before any gradient step. Training thus starts from the deep only baseline, and disabling fusion recovers that baseline exactly. The fused context feeds every cross-attention block, so the head sees semantics and spatial detail at all of its depths, at the cost of a single linear layer. We use a fixed intermediate index $m = 6$; Section~\ref{sec:shallow_fusion_ablation} compares the fused context against the deep only context.

\subsection{Flow Matching}

FabriVLA generates actions via flow matching \citep{lipman2023flow}, a generative framework that learns a time dependent vector field transporting samples from a simple base distribution to the target data distribution. In our setting, actions and states are min-max normalized per dimension to $[-1,1]$, so the base distribution is uniform noise on that cube and the target is the distribution of expert action chunks.

\subsubsection{Flow interpolation.} Let $\mathbf{a} \in \R^{H \times D}$ be a normalized padded expert action chunk, $\boldsymbol{\epsilon} \sim \mathcal{U}([-1,1]^{H \times D})$ be noise, and $t \sim \mathrm{Beta}(2,2)$. The interpolation is
\begin{equation}
    \mathbf{x}_t = (1-t)\,\boldsymbol{\epsilon} + t\,\mathbf{a}.
\end{equation}

\subsubsection{Velocity field prediction.} The action head predicts the velocity field $\mathbf{v}_\theta$ introduced above and minimizes
\begin{equation}
    \cL(\theta) = \E_{t, \boldsymbol{\epsilon}, \mathbf{a}} \left[ \| \mathbf{v}_\theta(\mathbf{x}_t, t \mid \cdot) - (\mathbf{a} - \boldsymbol{\epsilon}) \|^2 \right].
\end{equation}

\subsubsection{Inference.} Starting from $\mathbf{x}_0 \sim \mathcal{U}([-1,1]^{H \times D})$, we integrate the velocity field with $N$ uniform Euler steps of size $\Delta t = 1/N$. The resulting chunk is denormalized, its $d$ valid embodiment dimensions are extracted, and its first $H_e$ steps are executed before the next query.

\begin{figure*}[t]
\centering
\includegraphics[width=0.98\textwidth]{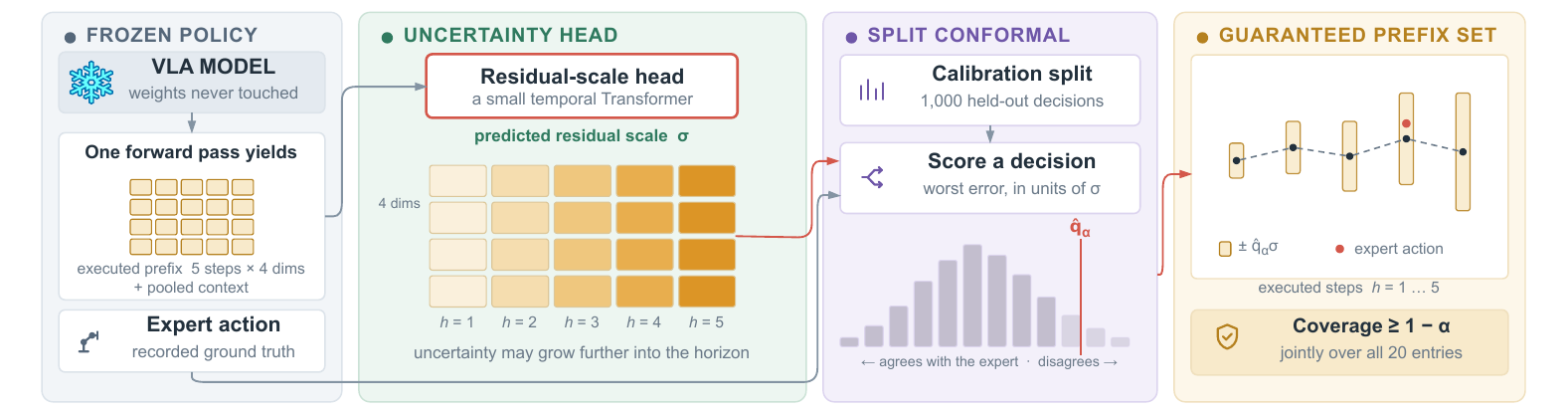}
\caption{Joint conformal action chunk calibration. The policy stays frozen: one query yields the executed prefix, pooled context, and robot state. A residual scale head (red outline) turns these into an elementwise scale $\sigma_\phi$; each calibration decision reduces to one score in units of $\sigma_\phi$, and one scalar quantile $\hat{q}_\alpha$ then covers all $H_e \times d$ executed entries jointly.}
\label{fig:jcac_pipeline}
\end{figure*}

% ============================================================================
% 4. JOINT CONFORMAL ACTION CHUNK CALIBRATION
% ============================================================================
\section{Joint Conformal Action Chunk Calibration}

JCAC answers one question at deployment time: before the prefix $\widehat{\mathbf{A}}^e_k$ leaves the controller, how far can it be from what an expert would have done? It returns a set that covers all $H_e \times d$ executed entries simultaneously and carries a distribution-free coverage guarantee, using only the policy query that produced the chunk. The base policy stays frozen. Figure~\ref{fig:jcac_pipeline} summarizes the pipeline.

\subsection{Conditional Residual Scale Head}

At replanning step $k$ we retain the executed prefix $\widehat{\mathbf{A}}^e_k$ and pool the fused VLM context that produced it,
\begin{equation}
\mathbf{c}_k=\operatorname{MeanPool}(\mathbf{C}_k),\qquad
\mathbf{g}_k=f_c([\mathbf{c}_k;\widetilde{\mathbf{s}}_k]),
\end{equation}
where $f_c$ encodes the visual-language context and the normalized padded state $\widetilde{\mathbf{s}}_k$ into a single condition token. Each predicted action step $h$ becomes a token $\mathbf{u}_{k,h}$ through a learned linear map and a temporal embedding, and the condition token emits FiLM scale and shift parameters that modulate every action token into $\widetilde{\mathbf{u}}_{k,h}$ (Appendix~B). A bidirectional temporal Transformer then jointly processes the condition token and all $H_e$ action tokens, and predicts positive elementwise residual scales
\begin{equation}
\boldsymbol{\sigma}_\phi
=\operatorname{softplus}\!\left(
h_\phi([\mathbf{g}_k;\widetilde{\mathbf{u}}_{k,1:H_e}])
\right)+\sigma_{\min}
\in\R_+^{H_e\times d},
\end{equation}
where $\sigma_{\min}>0$ is a fixed floor that keeps the normalized calibration scores defined below finite. Treating the prefix jointly rather than step by step lets the head express uncertainty that accumulates along the chunk. Because it reuses the context already computed by the single FabriVLA query, it costs no additional policy forward pass.

The base policy remains frozen. We query it at states from newly collected expert demonstrations that are held out from FabriVLA training. For each selected expert state, the next $H_e$ recorded expert actions form the reference prefix $\mathbf{A}^{*}_k$. The residual head is fitted on a split disjoint from calibration using a Gaussian negative log-likelihood on the residuals,
\begin{equation}
\cL_{\mathrm{scale}}=
\frac{1}{H_ed}\sum_{h=1}^{H_e}\sum_{j=1}^{d}
\left[
\frac{(A^{*}_{k,h,j}-\widehat{A}^{e}_{k,h,j})^2}
{2\sigma_{\phi,h,j}^2}+\log\sigma_{\phi,h,j}
\right].
\end{equation}
The Gaussian form is a convenience for fitting the scale. It governs how tight the final set is, not whether the set is valid, since conformal validity does not require Gaussian residuals.

\subsection{Split Conformal Calibration}

For calibration sample $i$, we use a single score for all executed steps and dimensions,
\begin{equation}
S_i=\max_{h\le H_e,\,j\le d}
\frac{|A^{*}_{i,h,j}-\widehat{A}^{e}_{i,h,j}|}
{\sigma_{\phi,i,h,j}}.
\end{equation}
The maximum is what makes the resulting set joint rather than pointwise: one quantile of $S$ controls every executed entry at once, with no union bound over the $H_e \times d$ entries. For $n$ calibration examples and target miscoverage $\alpha$, let $\widehat q_\alpha$ be the $\lceil(n+1)(1-\alpha)\rceil$-th smallest score, which requires $n \ge \lceil 1/\alpha \rceil - 1$ so that this rank exists. JCAC returns the joint prediction set
\begin{equation}
\mathcal{C}_\alpha(\mathbf{o}_k,\mathbf{s}_k,\ell)=
\left\{\mathbf{A}: |A_{h,j}-\widehat A^e_{h,j}|
\le\widehat q_\alpha\sigma_{\phi,h,j},\ \forall h,j\right\}.
\end{equation}
Under exchangeability of calibration and test examples, this construction provides marginal joint prefix coverage at least $1-\alpha$ \citep{vovk2005algorithmic,angelopoulos2021conformal}. We call the per-element half width $\widehat q_\alpha\sigma_{\phi,h,j}$ that entry's \emph{radius}.

The guarantee concerns agreement with expert actions on exchangeable data, so results obtained outside that setting, including the closed loop and cross-policy studies below, are reported as ranking measurements.

The calibrated radii expose a structured uncertainty signal to downstream monitoring or replanning rules. Section~\ref{sec:reliability_eval} instantiates both: a retrospective selective deferral curve over episodes, and a per-decision replanning trigger driven by a frozen radius threshold.

\subsection{Training}

\subsubsection{Policy training.} FabriVLA is trained in a single stage: the backbone starts from its public pretrained checkpoint with no robot data pretraining. The same recipe trains both the Meta-World and the real robot policy; the Meta-World run uses five NVIDIA RTX~PRO~6000 Blackwell GPUs. Corpora, hyperparameters, and the mixed precision requirement for joint backbone updates are given in Appendix~A.

\subsubsection{Uncertainty head training.} JCAC never updates the policy. With FabriVLA frozen, the residual scale head is the only trainable component, fit on a single NVIDIA GeForce RTX~5090 by minimizing a Gaussian negative log-likelihood on the residual between the executed prefix and the corresponding expert actions, using demonstrations disjoint from policy training. Conformal calibration then consumes a further disjoint split and takes no gradient step. Splits and head dimensions are given in Appendix~B.

% ============================================================================
% 5. EXPERIMENTS
% ============================================================================
\section{Experiments}

We design our experiments to address the following key Research Questions (RQs):
\begin{itemize}[leftmargin=*,itemsep=1pt,topsep=3pt]
\item RQ1: Can a lightweight VLA trained in a single stage, without robot data pretraining, match larger VLAs on multi-task manipulation?
\item RQ2: Does JCAC turn a single query to a frozen policy into a calibrated and informatively tight uncertainty set over the executed action prefix?
\item RQ3: Does the combined system run on a real robot at closed-loop rates, and does a deployment time score computed from the same frozen query separate a nominal layout from a shifted one?
\end{itemize}

\subsection{Setup}

We evaluate in three settings. \textbf{Meta-World MT50} \citep{yu2020metaworld} comprises 50 distinct robotic manipulation tasks ranging from simple reaching and pushing to complex articulated object manipulation (e.g., door opening, peg insertion, nut assembly); it carries both the policy comparison and the JCAC calibration study. \textbf{LIBERO-Safety} \citep{cui2026liberosafety}, built on LIBERO \citep{liu2023libero}, provides physical safety task configurations in which constraint violations are annotated separately from task success; there JCAC is attached to the released $\pi_{0.5}$ \citep{intelligence2025pi05} checkpoint distributed with the benchmark \citep{cui2026liberosafety}. Finally, we deploy FabriVLA on a Unitree D1-T arm performing cube-to-tray placement from two RGB views, where path adjacent obstacles probe behavior under real sensing and control.

\subsubsection{Model configuration.} FabriVLA pairs the pretrained InternVL3.5-1B backbone, of which the first $M = 14$ language-model layers are retained and whose final layer is fused with layer $m = 6$, with an $L = 8$ block flow matching head predicting chunks of $H = 50$ steps. Meta-World actuates $d = 4$ of the $D = 24$ padded action dimensions. The remaining architecture and training details are given in Appendix~A.

\subsection{Multi-Task Manipulation Performance}

To address RQ1, we compare FabriVLA against recent VLAs on MT50. We evaluate all 50 tasks over 10 episodes each. Rollouts use an execution horizon of five steps, written H5 below, and $N = 50$ integration steps; an episode succeeds if the simulator raises its success flag at any step within a 400 step horizon. The reported \emph{average} success rate is the mean of the four difficulty tier scores. Baseline averages are computed by the same formula from the per tier scores reported in their papers; Appendix~A gives the full metric definition.

\subsubsection{Comparison with recent VLAs.} Table~\ref{tab:comparison} places FabriVLA against recent VLA models on Meta-World MT50. Despite using only 0.88B parameters and no robot data pretraining, FabriVLA achieves the best average success rate (90.0\%) among the compared methods, ahead of LA4VLA (87.5\%) and the depth augmented Evo-Depth (84.4\%). It is strongest on the easy tier and second strongest on the medium, hard, and very hard tiers, while LA4VLA reports the highest medium and very hard scores. Overall, the comparison shows that FabriVLA attains competitive or better manipulation performance with a compact model.
\begin{table*}[t]
\centering
\caption{Comparison on Meta-World MT50 against recent VLA models. ``Robo-Pre.'' is the reported robot data pretraining recipe. \textbf{Bold} and \underline{underline} mark the best and second best per column, and the smallest and second smallest under ``Params''. Baseline numbers are from the corresponding papers.}
\label{tab:comparison}
\setlength{\tabcolsep}{6pt}
\begin{tabular}{lccccccc}
\toprule
\textbf{Model} & \textbf{Params} & \textbf{Robo-Pre.} & \textbf{Easy} & \textbf{Med.} & \textbf{Hard} & \textbf{V.Hard} & \textbf{Avg.} \\
\midrule
TinyVLA~\citep{wen2024tinyvla}    & 1.3B & No  & 77.6 & 21.5 & 11.4 & 15.8 & 31.6 \\
$\pi_0$~\citep{black2024pi0}      & 3.5B & Yes & 71.8 & 48.2 & 41.7 & 30.0 & 47.9 \\
SmolVLA~\citep{shukor2025smolvla} & 2.3B & No  & 87.1 & 51.8 & 70.0 & 64.0 & 68.2 \\
RoboTron-Mani~\citep{yan2025robotron} & 4B & No & 85.5 & 67.7 & 76.7 & 81.0 & 77.7 \\
Evo-1~\citep{lin2025evo1}         & \textbf{0.8B} & No & \underline{89.2} & 76.8 & 77.2 & 79.2 & 80.6 \\
Evo-Depth~\citep{evodepth2026}    & 0.9B & No & 83.1 & 84.7 & \textbf{87.3} & 82.4 & 84.4 \\
LA4VLA~\citep{lin2026la4vla}      & 1B & MixPT & 88.9 & \textbf{94.5} & 66.7 & \textbf{100.0} & \underline{87.5} \\
\midrule
\textbf{FabriVLA (Ours)} & \underline{0.88B} & No & \textbf{95.0} & \underline{88.2} & \underline{86.7} & \underline{90.0} & \textbf{90.0} \\
\bottomrule
\end{tabular}
\end{table*}

\subsection{Joint Calibration and Closed Loop Reliability}
\label{sec:reliability_eval}

To address RQ2, we evaluate JCAC on independent expert state splits, where coverage is the quantity being tested, then in closed loop and across policies, where the measurement is ranking. Every design choice is frozen on a development corpus; a separate corpus under disjoint seeds then supplies 1{,}000 final calibration and 750 disjoint final test decisions, one per episode~(Appendix~B). We estimate $\widehat q_\alpha$ at $\alpha=0.1$ on the calibration split and score joint prefix coverage and radius once on the held out decisions.

Table~\ref{tab:jcac_final} compares JCAC against the natural baseline, a single global conformal radius fit on the same split. Conditioning the radius on the query is what buys tightness: the baseline reaches nominal coverage by assigning every action element the same wide radius of 1.606, giving 3.3 times JCAC's mean radius of 0.480, at a 0.901 radius error correlation. The two are not compared at identical coverage: the baseline records 89.33\% and JCAC 88.80\% against the 90\% target, the latter inside its 95\% Wilson interval of 86.3--90.9\%, so JCAC is the marginally looser of the two in coverage while being the tighter in width. Conditioning also redistributes width rather than removing it uniformly, which is why the two statistics in Table~\ref{tab:jcac_final} move in opposite directions: JCAC's mean radius is the smallest in the table, but its P90 and P95 of the \emph{per-decision maximum} radius exceed the baseline's flat 1.606. On 65.7\% of the 750 test decisions the entire executed prefix is bounded more tightly than the global constant allows, and the remaining 34.3\% absorb the width the global radius had spread evenly over every decision (Appendix~C).

\begin{table}[t]
\centering
\small
\caption{Final JCAC evaluation on 750 held out decisions at a 90\% joint prefix target. Cov.\ is the empirical joint prefix coverage, the fraction of test decisions whose $H_e \times d$ executed entries all fall inside $\mathcal{C}_\alpha$. The width columns summarize two different quantities and are not comparable to each other: ``Mean rad.'' averages the radius over all valid action elements, whereas ``P90 max'' and ``P95 max'' are percentiles of the \emph{per-decision maximum} radius. Corr.\ is the correlation between the per-decision maximum radius and the per-decision maximum residual.}
\label{tab:jcac_final}
\setlength{\tabcolsep}{4pt}
\begin{tabular}{lrrrrr}
\toprule
\textbf{Method} & \textbf{Cov.} & \textbf{Mean} & \textbf{P90} & \textbf{P95} & \textbf{Corr.} \\
\midrule
Global radius & 89.33\% & 1.606 & 1.606 & 1.606 & -- \\
MLP scale head & 90.67\% & 0.592 & 4.285 & 5.713 & 0.812 \\
JCAC (ours) & 88.80\% & \textbf{0.480} & 3.710 & 4.091 & \textbf{0.901} \\
\bottomrule
\end{tabular}
\end{table}

We then run the calibrated score in closed loop on 500 clean MT50 rollouts under fixed-H5 execution, where the episode maximum radius ranks eventual failure at 0.664 AUROC (95\% episode bootstrap CI 0.577--0.745). The score is used here as a ranking signal over episodes: replanning rules driven by the frozen calibration threshold leave clean MT50 success at its fixed-H5 level, so the score is reported without a control rule attached. Thresholds, horizon control, and per-decision statistics are given in Appendix~C.

We repeat the same fixed H5 measurement under five predeclared perturbations (brightness, state noise, one step action delay, 0.8 XYZ action scaling, and controlled mid episode object displacement) with model, seeds, and threshold held fixed. Coverage on held out expert states stays near its calibrated level throughout, and episode level failure AUROC tracks condition severity, rising to 0.747 (0.619--0.861) under object displacement, which is by far the most severe condition: it drops success from 91.0\% to 15.0\%, against at most 4.0 points for the other four. The full stress test table is given in Appendix~C.

\subsubsection{Cross policy confirmation on LIBERO-Safety.}
We next attach the same JCAC construction to the frozen official $\pi_{0.5}$ LIBERO-Safety checkpoint \citep{cui2026liberosafety}. The residual head uses the checkpoint's PaliGemma context and is fit on the official collision free expert corpus without modifying the base policy. Because this corpus was also used to fine-tune the base policy, we report ranking rather than coverage here. After a frozen development gate (Appendix~C), the confirmatory run logs 600 episodes on disjoint initial states under fixed H5 execution, containing 435 successes and 165 non-success episodes. We score non-success as positive and abbreviate it as \emph{task failure} below, so the no-deferral prevalence is 27.5\%; constraint events are treated on their own in the event aligned study that follows.

\begin{figure}[t]
\centering
\includegraphics[width=\columnwidth]{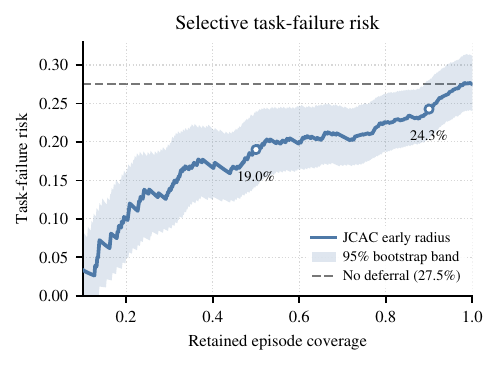}
\caption{Selective task failure risk over the 600 confirmatory LIBERO-Safety episodes, retaining episodes from low to high first five JCAC radius. Shading is a 95\% episode bootstrap band; the dashed line is the 27.5\% no deferral prevalence. The curve measures ranking only; no deferral is executed.}
\label{fig:jcac_reliability}
\end{figure}

The predeclared primary score is the maximum JCAC radius in the first five decisions, which ranks eventual task failure at 0.674 AUROC (95\% bootstrap CI 0.626--0.721). As a retrospective selective risk diagnostic, Figure~\ref{fig:jcac_reliability} shows that retaining the lowest radius half of the episodes lowers observed task failure risk from the 27.5\% no deferral prevalence to 19.0\%. The ranking also survives conditioning on task identity: pair weighted success failure concordance within configurations containing both outcomes is 0.616.

Constraint events are event local rather than episode level, so we treat the first collision aligned study as development, select its 10 step sensitivity window, and freeze H10 before a targeted confirmation on 390 previously unused initial state slots, giving 780 repeated measure rollouts separate from the 600 episodes above, with time and configuration matched controls. That run records 31 constraint events, of which 29 have valid pre event windows and matched controls; H10 reaches 0.782 AUROC (0.699--0.853), ranks ahead of the longer H25 and H50 windows, and passes the frozen event count, AUROC, interval, and direction criteria. Both LIBERO-Safety results are ranking measurements: the score is reported as a continuous prioritization signal rather than a binary alarm, and no deferral or collision intervention is executed. Operating point statistics and the confirmation protocol are given in Appendix~C.

\subsection{Ablation Study}
\subsubsection{Component Ablation.}
\label{sec:shallow_fusion_ablation}
We ablate the two architectural components at 80k training steps, evaluating every variant over the same 500 initial conditions. Removing gated self-attention lowers average success from 81.5\% to 76.2\%, with the loss concentrated in the hard and very hard tiers, which fall from 75.0\% to 66.7\% and from 78.0\% to 68.0\%; removing shallow fusion lowers average success to 77.9\%. Both components thus contribute to the full model. These runs use a shorter schedule and a longer execution horizon than the released checkpoint, so their absolute rates are not comparable with Table~\ref{tab:comparison}. Per tier scores, paired statistics, training recipes, and the remaining head variants are given in Appendix~A.

\subsubsection{JCAC Scale-Head Ablation.}
The last two rows of Table~\ref{tab:jcac_final} isolate the residual head architecture on exactly the same final splits. Both sit close to the 90\% target, as conformal calibration enforces by construction; the head only decides how tightly that coverage is distributed across decisions. Replacing the MLP with the development selected temporal Transformer reduces mean radius by 18.9\% and P95 maximum radius by 28.4\%, and raises the radius error correlation from 0.812 to 0.901, at roughly twice the head parameter count.

\subsection{Real Robot Deployment}
\label{sec:real_robot}

\begin{figure}[t]
\centering
\includegraphics[width=0.85\columnwidth]{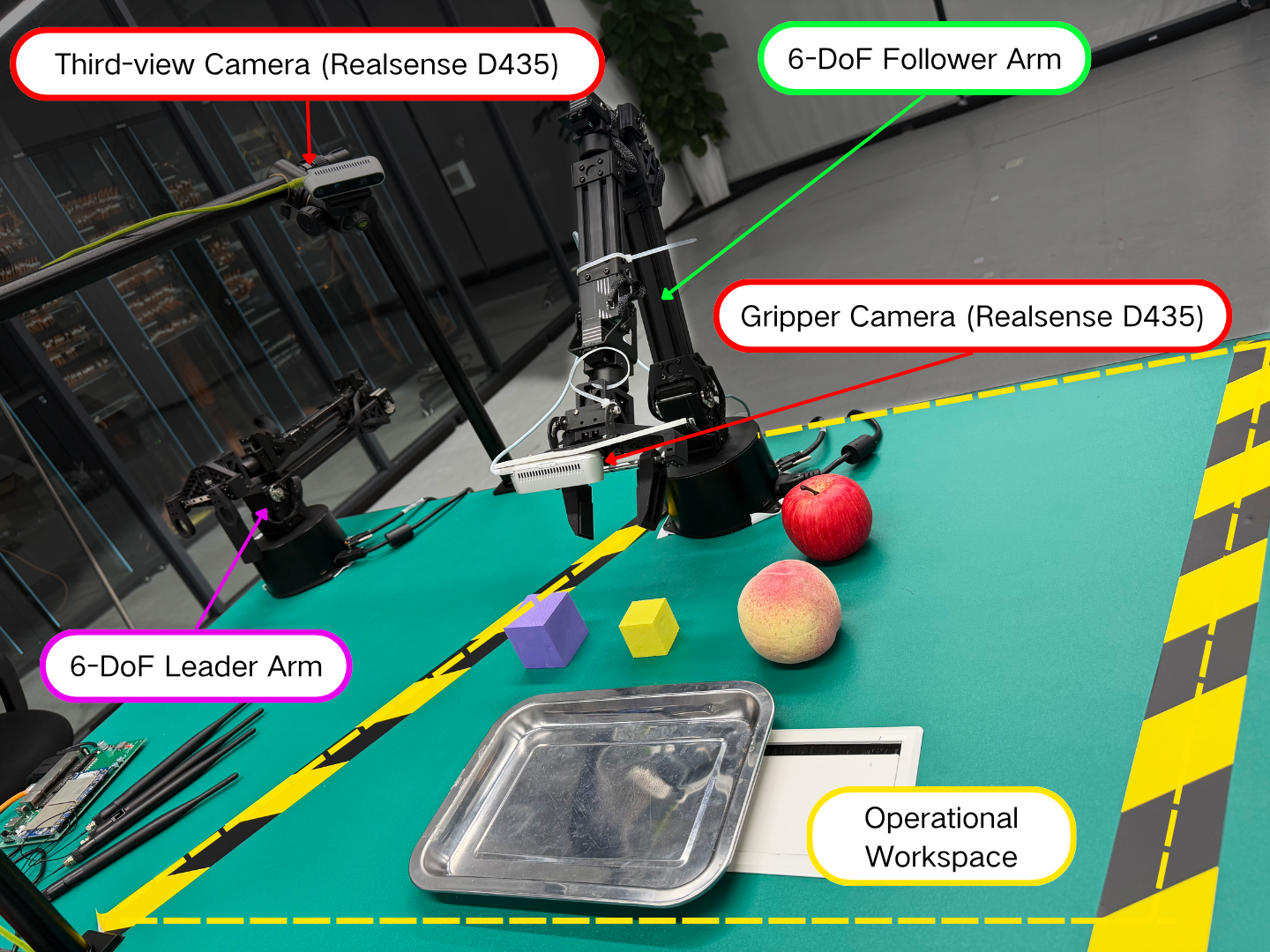}
\caption{Real robot setup: the D1-T follower arm performs cube-to-tray placement from a fixed third person and a gripper proximal RGB view, with the leader used only for teleoperated data collection (Appendices~A and~D).}
\label{fig:real_robot_setup}
\end{figure}

To address RQ3, we train a separate real robot policy from 50 demonstrations of cube-to-tray placement on the D1-T platform of Figure~\ref{fig:real_robot_setup}, and fit and calibrate JCAC on 15 disjoint episodes each. The policy consumes both RGB views and the seven joint dimensions, predicts 50 step chunks, and executes 10 before replanning. All reported runs freeze the policy, calibration, cameras, and controller (Appendix~D).

Beyond JCAC's action prefix set, we evaluate a complementary deployment diagnostic on the frozen VLM context: the mean cosine distance to the five nearest contexts of the residual fit split, in the spirit of feature space novelty detection~\citep{sun2022knnood,farid2022taskdriven}, thresholded at a finite sample 90\% conformal quantile from the disjoint calibration split. We compare 10 clean center trials against 10 trials with a fixed path adjacent obstacle under matched nighttime illumination. The study is logging only: no score changes an executed action and no operator intervenes. Table~\ref{tab:d1t_deployment} reports the outcome.

The continuous score provides evidence for context shift and execution risk prioritization, though it is not interpreted as an obstacle detector or failure probability. Per-trial results and the binary rule analysis are given in Appendix~D.

\begin{table}[t]
\centering
\small
\caption{Frozen D1-T deployment study, 10 trials per layout. ``Mean'' averages the episode mean context score, and $>\tau$ counts decision scores above the fixed threshold $\tau=0.0393$. The episode mean score separates the two layouts at 0.970 AUROC (stratified episode bootstrap 95\% CI 0.880--1.000). The diagnostic is logging only.}
\label{tab:d1t_deployment}
\begin{tabular*}{\columnwidth}{@{\extracolsep{\fill}}lrrrr@{}}
\toprule
\textbf{Condition} & \textbf{$N$} & \textbf{Success} & \textbf{Mean} & \textbf{$>\tau$} \\
\midrule
Clean (ID) & 10 & 10/10 & 0.02878 & 34/223 \\
Obstacle (OOD) & 10 & 1/10 & 0.03795 & 318/738 \\
\bottomrule
\end{tabular*}
\end{table}

\subsubsection{On robot inference latency.}
On an NVIDIA GeForce RTX~5090 workstation, the complete dual view policy and uncertainty query takes a median 220.5\,ms and P90 225.7\,ms; the JCAC head contributes 0.481\,ms of that at batch one, for 0.58M added parameters. A 10 step prefix therefore costs one query per second of motion, and asynchronous inference overlaps that computation with execution of the preceding chunk.

% ============================================================================
% 6. CONCLUSION
% ============================================================================
\section{Conclusion}

We presented FabriVLA, a lightweight VLA that reaches 90.0\% average success on Meta-World MT50 with 0.88B parameters and single stage training, and whose recipe carries over unchanged to a real robot arm. JCAC then makes such a policy auditable without retraining it: one query returns a joint error set over the action prefix that is actually executed, for 0.58M added parameters and sub-millisecond head latency. Its radii rank task failure and constraint events on LIBERO-Safety, and a companion context score separates a layout shift in real deployment. These are ranking signals rather than alarms, so downstream safety decisions require separate closed loop validation. Together they show that a compact VLA can be both strong and auditable for the cost of one forward pass.

% ============================================================================
% BIBLIOGRAPHY
% ============================================================================
\bibliography{refs}

@InProceedings{pmlr-v229-zitkovich23a,
  title = 	 {RT-2: Vision-Language-Action Models Transfer Web Knowledge to Robotic Control},
  author =       {Zitkovich, Brianna and Yu, Tianhe and Xu, Sichun and Xu, Peng and Xiao, Ted and Xia, Fei and Wu, Jialin and Wohlhart, Paul and Welker, Stefan and Wahid, Ayzaan and Vuong, Quan and Vanhoucke, Vincent and Tran, Huong and Soricut, Radu and Singh, Anikait and Singh, Jaspiar and Sermanet, Pierre and Sanketi, Pannag R. and Salazar, Grecia and Ryoo, Michael S. and Reymann, Krista and Rao, Kanishka and Pertsch, Karl and Mordatch, Igor and Michalewski, Henryk and Lu, Yao and Levine, Sergey and Lee, Lisa and Lee, Tsang-Wei Edward and Leal, Isabel and Kuang, Yuheng and Kalashnikov, Dmitry and Julian, Ryan and Joshi, Nikhil J. and Irpan, Alex and Ichter, Brian and Hsu, Jasmine and Herzog, Alexander and Hausman, Karol and Gopalakrishnan, Keerthana and Fu, Chuyuan and Florence, Pete and Finn, Chelsea and Dubey, Kumar Avinava and Driess, Danny and Ding, Tianli and Choromanski, Krzysztof Marcin and Chen, Xi and Chebotar, Yevgen and Carbajal, Justice and Brown, Noah and Brohan, Anthony and Arenas, Montserrat Gonzalez and Han, Kehang},
  booktitle = 	 {Proceedings of the Conference on Robot Learning},
  pages = 	 {2165--2183},
  year = 	 {2023},
  editor = 	 {Tan, Jie and Toussaint, Marc and Darvish, Kourosh},
  volume = 	 {229},
  series = 	 {Proceedings of Machine Learning Research},
  month = 	 {06--09 Nov},
  publisher =    {PMLR},
  url = 	 {https://proceedings.mlr.press/v229/zitkovich23a.html}
}

@misc{kim2024openvla,
      title={OpenVLA: An Open-Source Vision-Language-Action Model}, 
      author={Moo Jin Kim and Karl Pertsch and Siddharth Karamcheti and Ted Xiao and Ashwin Balakrishna and Suraj Nair and Rafael Rafailov and Ethan Foster and Grace Lam and Pannag Sanketi and Quan Vuong and Thomas Kollar and Benjamin Burchfiel and Russ Tedrake and Dorsa Sadigh and Sergey Levine and Percy Liang and Chelsea Finn},
      year={2024},
      eprint={2406.09246},
      archivePrefix={arXiv},
      primaryClass={cs.RO},
      url={https://arxiv.org/abs/2406.09246}, 
}

@misc{black2024pi0,
      title={$\pi_0$: A Vision-Language-Action Flow Model for General Robot Control}, 
      author={Kevin Black and Noah Brown and Danny Driess and Adnan Esmail and Michael Equi and Chelsea Finn and Niccolo Fusai and Lachy Groom and Karol Hausman and Brian Ichter and Szymon Jakubczak and Tim Jones and Liyiming Ke and Sergey Levine and Adrian Li-Bell and Mohith Mothukuri and Suraj Nair and Karl Pertsch and Lucy Xiaoyang Shi and James Tanner and Quan Vuong and Anna Walling and Haohuan Wang and Ury Zhilinsky},
      year={2024},
      eprint={2410.24164},
      archivePrefix={arXiv},
      primaryClass={cs.LG},
      url={https://arxiv.org/abs/2410.24164}, 
}

@InProceedings{lin2025evo1,
    author    = {Lin, Tao and Zhong, Yilei and Du, Yuxin and Zhang, Jingjing and Liu, Jiting and Chen, Yinxinyu and Gu, Encheng and Liu, Ziyan and Cai, Hongyi and Zou, Yanwen and Zou, Lixing and Zhou, Zhaoye and Li, Gen and Zhao, Bo},
    title     = {Evo-1: Lightweight Vision-Language-Action Model with Preserved Semantic Alignment},
    booktitle = {Proceedings of the IEEE/CVF Conference on Computer Vision and Pattern Recognition},
    month     = {June},
    year      = {2026},
    pages     = {13397-13406}
}

@misc{wang2025internvl35,
      title={InternVL3.5: Advancing Open-Source Multimodal Models in Versatility, Reasoning, and Efficiency}, 
      author={Weiyun Wang and Zhangwei Gao and Lixin Gu and Hengjun Pu and Long Cui and Xingguang Wei and Zhaoyang Liu and Linglin Jing and Shenglong Ye and Jie Shao and Zhaokai Wang and Zhe Chen and Hongjie Zhang and Ganlin Yang and Haomin Wang and Qi Wei and Jinhui Yin and Wenhao Li and Erfei Cui and Guanzhou Chen and Zichen Ding and Changyao Tian and Zhenyu Wu and Jingjing Xie and Zehao Li and Bowen Yang and Yuchen Duan and Xuehui Wang and Zhi Hou and Haoran Hao and Tianyi Zhang and Songze Li and Xiangyu Zhao and Haodong Duan and Nianchen Deng and Bin Fu and Yinan He and Yi Wang and Conghui He and Botian Shi and Junjun He and Yingtong Xiong and Han Lv and Lijun Wu and Wenqi Shao and Kaipeng Zhang and Huipeng Deng and Biqing Qi and Jiaye Ge and Qipeng Guo and Wenwei Zhang and Songyang Zhang and Maosong Cao and Junyao Lin and Kexian Tang and Jianfei Gao and Haian Huang and Yuzhe Gu and Chengqi Lyu and Huanze Tang and Rui Wang and Haijun Lv and Wanli Ouyang and Limin Wang and Min Dou and Xizhou Zhu and Tong Lu and Dahua Lin and Jifeng Dai and Weijie Su and Bowen Zhou and Kai Chen and Yu Qiao and Wenhai Wang and Gen Luo},
      year={2025},
      eprint={2508.18265},
      archivePrefix={arXiv},
      primaryClass={cs.CV},
      url={https://arxiv.org/abs/2508.18265}, 
}

@misc{lipman2023flow,
      title={Flow Matching for Generative Modeling}, 
      author={Yaron Lipman and Ricky T. Q. Chen and Heli Ben-Hamu and Maximilian Nickel and Matt Le},
      year={2023},
      eprint={2210.02747},
      archivePrefix={arXiv},
      primaryClass={cs.LG},
      url={https://arxiv.org/abs/2210.02747}, 
}

@article{chi2023diffusion,
  title={Diffusion Policy: Visuomotor Policy Learning via Action Diffusion},
  author={Chi, Cheng and Xu, Zhenjia and Feng, Siyuan and Cousineau, Eric and Du, Yilun and Burchfiel, Benjamin and Tedrake, Russ and Song, Shuran},
  journal={The International Journal of Robotics Research},
  volume={44},
  number={10-11},
  pages={1684--1704},
  year={2025},
  publisher={Sage Publications Sage UK: London, England}
}

@InProceedings{yu2020metaworld,
  title = 	 {Meta-World: A Benchmark and Evaluation for Multi-Task and Meta Reinforcement Learning},
  author =       {Yu, Tianhe and Quillen, Deirdre and He, Zhanpeng and Julian, Ryan and Hausman, Karol and Finn, Chelsea and Levine, Sergey},
  booktitle = 	 {Proceedings of the Conference on Robot Learning},
  pages = 	 {1094--1100},
  year = 	 {2020},
  editor = 	 {Kaelbling, Leslie Pack and Kragic, Danica and Sugiura, Komei},
  volume = 	 {100},
  series = 	 {Proceedings of Machine Learning Research},
  month = 	 {30 Oct--01 Nov},
  publisher =    {PMLR},
  url = 	 {https://proceedings.mlr.press/v100/yu20a.html},
}

@ARTICLE{wen2024tinyvla,
  author={Wen, Junjie and Zhu, Yichen and Li, Jinming and Zhu, Minjie and Tang, Zhibin and Wu, Kun and Xu, Zhiyuan and Liu, Ning and Cheng, Ran and Shen, Chaomin and Peng, Yaxin and Feng, Feifei and Tang, Jian},
  journal={IEEE Robotics and Automation Letters}, 
  title={TinyVLA: Toward Fast, Data-Efficient Vision-Language-Action Models for Robotic Manipulation}, 
  year={2025},
  volume={10},
  number={4},
  pages={3988-3995},
  doi={10.1109/LRA.2025.3544909}}

@misc{shukor2025smolvla,
      title={SmolVLA: A Vision-Language-Action Model for Affordable and Efficient Robotics}, 
      author={Mustafa Shukor and Dana Aubakirova and Francesco Capuano and Pepijn Kooijmans and Steven Palma and Adil Zouitine and Michel Aractingi and Caroline Pascal and Martino Russi and Andres Marafioti and Simon Alibert and Matthieu Cord and Thomas Wolf and Remi Cadene},
      year={2025},
      eprint={2506.01844},
      archivePrefix={arXiv},
      primaryClass={cs.LG},
      url={https://arxiv.org/abs/2506.01844}, 
}

@inproceedings{yan2025robotron,
  title={RoboTron-Mani: All-in-One Multimodal Large Model for Robotic Manipulation},
  author={Yan, Feng and Liu, Fanfan and Huang, Yiyang and Guan, Zechao and Zheng, Liming and Zhong, Yufeng and Feng, Chengjian and Ma, Lin},
  booktitle={Proceedings of the IEEE/CVF International Conference on Computer Vision},
  pages={13707--13718},
  year={2025}
}

@misc{evodepth2026,
      title={Evo-Depth: A Lightweight Depth-Enhanced Vision-Language-Action Model}, 
      author={Tao Lin and Yuxin Du and Jiting Liu and Nuobei Zhu and Yunhe Li and Yuqian Fu and Yinxinyu Chen and Hongyi Cai and Zewei Ye and Bing Cheng and Kai Ye and Yiran Mao and Yilei Zhong and MingKang Dong and Junchi Yan and Gen Li and Bo Zhao},
      year={2026},
      eprint={2605.14950},
      archivePrefix={arXiv},
      primaryClass={cs.CV},
      url={https://arxiv.org/abs/2605.14950}, 
}

@misc{lin2026la4vla,
      title={LA4VLA: Learning to Act without Seeing via Language-Action Pretraining}, 
      author={Tao Lin and Yuxin Du and Yiran Mao and Zewei Ye and Yilei Zhong and Bing Cheng and Yiming Wang and Jiting Liu and Yang Tian and Junchi Yan and Feiran Wu and Zenan Meng and Hu Wei and Yuqian Fu and Gen Li and Bo Zhao},
      year={2026},
      eprint={2606.27295},
      archivePrefix={arXiv},
      primaryClass={cs.RO},
      url={https://arxiv.org/abs/2606.27295}, 
}

@misc{angelopoulos2021conformal,
      title={A Gentle Introduction to Conformal Prediction and Distribution-Free Uncertainty Quantification},
      author={Anastasios N. Angelopoulos and Stephen Bates},
      year={2021},
      eprint={2107.07511},
      archivePrefix={arXiv},
      primaryClass={cs.LG},
      url={https://arxiv.org/abs/2107.07511},
}

@inproceedings{romano2019cqr,
      title={Conformalized Quantile Regression},
      author={Yaniv Romano and Evan Patterson and Emmanuel J. Cand\`es},
      booktitle={Advances in Neural Information Processing Systems},
      year={2019},
}

@misc{liu2023libero,
      title={LIBERO: Benchmarking Knowledge Transfer for Lifelong Robot Learning},
      author={Bo Liu and Yifeng Zhu and Chongkai Gao and Yihao Feng and Qiang Liu and Yuke Zhu and Peter Stone},
      year={2023},
      eprint={2306.03310},
      archivePrefix={arXiv},
      primaryClass={cs.RO},
      url={https://arxiv.org/abs/2306.03310},
}

@misc{cui2026liberosafety,
      title={LIBERO-Safety: A Comprehensive Benchmark for Physical and Semantic Safety in Vision-Language-Action Models},
      author={Rongxu Cui and Zongzheng Zhang and Jingrui Pang and Haohan Chi and Jinbang Guo and Saining Zhang and Shaoxuan Xie and Xin Jin and Yao Mu and Jiaolong Yang and Guocai Yao and Xianyuan Zhan and Ya-Qin Zhang and Hao Zhao},
      year={2026},
      eprint={2606.23686},
      archivePrefix={arXiv},
      primaryClass={cs.RO},
      url={https://arxiv.org/abs/2606.23686},
}

@book{vovk2005algorithmic,
      title={Algorithmic Learning in a Random World},
      author={Vovk, Vladimir and Gammerman, Alexander and Shafer, Glenn},
      year={2005},
      publisher={Springer},
}

@inproceedings{ren2023knowno,
      title={Robots That Ask For Help: Uncertainty Alignment for Large Language Model Planners},
      author={Ren, Allen Z. and Dixit, Anushri and Bodrova, Alexandra and Singh, Sumeet and Tu, Stephen and Brown, Noah and Xu, Peng and Takayama, Leila and Xia, Fei and Varley, Jake and Xu, Zhenjia and Sadigh, Dorsa and Zeng, Andy and Majumdar, Anirudha},
      booktitle={Conference on Robot Learning (CoRL)},
      year={2023},
}

@article{lindemann2023safe,
      title={Safe Planning in Dynamic Environments Using Conformal Prediction},
      author={Lindemann, Lars and Cleaveland, Matthew and Shim, Gihyun and Pappas, George J.},
      journal={IEEE Robotics and Automation Letters},
      volume={8},
      number={8},
      pages={5116--5123},
      year={2023},
      publisher={IEEE},
}

@inproceedings{dixit2023adaptive,
      title={Adaptive Conformal Prediction for Motion Planning among Dynamic Agents},
      author={Dixit, Anushri and Lindemann, Lars and Wei, Skylar X. and Cleaveland, Matthew and Pappas, George J. and Burdick, Joel W.},
      booktitle={Learning for Dynamics and Control Conference (L4DC)},
      year={2023},
      organization={PMLR},
}

@inproceedings{sun2024copula,
      title={Copula Conformal Prediction for Multi-step Time Series Forecasting},
      author={Sun, Sophia Huiwen and Yu, Rose},
      booktitle={International Conference on Learning Representations (ICLR)},
      year={2024},
}

@article{chen2026reconvla,
      title={ReconVLA: An Uncertainty-Guided and Failure-Aware Vision-Language-Action Framework for Robotic Control},
      author={Chen, Lingling and Lyu, Zongyao and Beksi, William J.},
      journal={arXiv preprint arXiv:2604.16677},
      year={2026},
}

@inproceedings{hariharan2015hypercolumns,
  author    = {Hariharan, Bharath and Arbel{\'a}ez, Pablo and Girshick, Ross and Malik, Jitendra},
  title     = {Hypercolumns for Object Segmentation and Fine-Grained Localization},
  booktitle = {Proceedings of the IEEE Conference on Computer Vision and Pattern Recognition},
  pages     = {447--456},
  year      = {2015}
}

@inproceedings{lin2017fpn,
  author    = {Lin, Tsung-Yi and Doll{\'a}r, Piotr and Girshick, Ross and He, Kaiming and Hariharan, Bharath and Belongie, Serge},
  title     = {Feature Pyramid Networks for Object Detection},
  booktitle = {Proceedings of the IEEE Conference on Computer Vision and Pattern Recognition},
  pages     = {2117--2125},
  year      = {2017}
}

@inproceedings{ranftl2021vision,
  author    = {Ranftl, Ren{\'e} and Bochkovskiy, Alexey and Koltun, Vladlen},
  title     = {Vision Transformers for Dense Prediction},
  booktitle = {Proceedings of the IEEE/CVF International Conference on Computer Vision},
  pages     = {12179--12188},
  year      = {2021}
}

@inproceedings{menda2019ensembledagger,
  author    = {Menda, Kunal and Driggs-Campbell, Katherine and Kochenderfer, Mykel J.},
  title     = {{EnsembleDAgger}: A Bayesian Approach to Safe Imitation Learning},
  booktitle = {IEEE/RSJ International Conference on Intelligent Robots and Systems},
  pages     = {5041--5048},
  year      = {2019}
}

@misc{intelligence2025pi05,
      title={{$\pi_{0.5}$}: a Vision-Language-Action Model with Open-World Generalization},
      author={{Physical Intelligence} and Black, Kevin and Brown, Noah and Darpinian, James and Dhabalia, Karan and Driess, Danny and others},
      year={2025},
      eprint={2504.16054},
      archivePrefix={arXiv},
      primaryClass={cs.RO},
      url={https://arxiv.org/abs/2504.16054}
}

@inproceedings{zhao2023act,
  title     = {Learning Fine-Grained Bimanual Manipulation with Low-Cost Hardware},
  author    = {Zhao, Tony Z. and Kumar, Vikash and Levine, Sergey and Finn, Chelsea},
  booktitle = {Proceedings of Robotics: Science and Systems (RSS)},
  year      = {2023}
}

@inproceedings{sun2022knnood,
  title     = {Out-of-Distribution Detection with Deep Nearest Neighbors},
  author    = {Sun, Yiyou and Ming, Yifei and Zhu, Xiaojin and Li, Yixuan},
  booktitle = {Proceedings of the 39th International Conference on Machine Learning},
  series    = {Proceedings of Machine Learning Research},
  volume    = {162},
  year      = {2022},
  publisher = {PMLR}
}

@inproceedings{farid2022taskdriven,
  title     = {Task-Driven Out-of-Distribution Detection with Statistical Guarantees for Robot Learning},
  author    = {Farid, Alec and Veer, Sushant and Majumdar, Anirudha},
  booktitle = {Proceedings of the 5th Conference on Robot Learning},
  series    = {Proceedings of Machine Learning Research},
  volume    = {164},
  year      = {2022},
  publisher = {PMLR}
}

\end{document}